\pdfoutput=1

\documentclass[11pt]{article}

\usepackage{bera}
\usepackage{listings}
\usepackage{xcolor}

\colorlet{punct}{red!60!black}
\definecolor{background}{HTML}{EEEEEE}
\definecolor{delim}{RGB}{20,105,176}
\colorlet{numb}{magenta!60!black}
\usepackage{longtable}
\usepackage{supertabular}
\lstdefinelanguage{json}{
    basicstyle=\normalfont\ttfamily,
    numbers=left,
    numberstyle=\scriptsize,
    stepnumber=1,
    numbersep=8pt,
    showstringspaces=false,
    breaklines=true,
    frame=lines,
    backgroundcolor=\color{background},
    literate=
     *{0}{{{\color{numb}0}}}{1}
      {1}{{{\color{numb}1}}}{1}
      {2}{{{\color{numb}2}}}{1}
      {3}{{{\color{numb}3}}}{1}
      {4}{{{\color{numb}4}}}{1}
      {5}{{{\color{numb}5}}}{1}
      {6}{{{\color{numb}6}}}{1}
      {7}{{{\color{numb}7}}}{1}
      {8}{{{\color{numb}8}}}{1}
      {9}{{{\color{numb}9}}}{1}
      {:}{{{\color{punct}{:}}}}{1}
      {,}{{{\color{punct}{,}}}}{1}
      {\{}{{{\color{delim}{\{}}}}{1}
      {\}}{{{\color{delim}{\}}}}}{1}
      {[}{{{\color{delim}{[}}}}{1}
      {]}{{{\color{delim}{]}}}}{1},
}

\usepackage[final]{acl}

\usepackage{times}
\usepackage{latexsym}
\usepackage{booktabs}
\usepackage{multirow}

\usepackage{xcolor}
\usepackage{lipsum}

\usepackage[T1]{fontenc}

\usepackage[utf8]{inputenc}

\usepackage{microtype}

\usepackage{inconsolata}

\usepackage{graphicx}

\usepackage{amsmath}
\newcommand{\red}[1]{\textcolor{red}{#1}}

\mathcode`*=\string"8000
\begingroup
\catcode`*=\active
\xdef*{\noexpand\textup{\string*}}
\endgroup

\title{Bridging Modalities: Enhancing Cross-Modality Hate Speech Detection with Few-Shot In-Context Learning}

\author{
 \textbf{Ming Shan Hee\textsuperscript{1,}}\thanks{These authors contributed equally to this work.},
 \textbf{Aditi Kumaresan\textsuperscript{1, *}},
 \textbf{Roy Ka-Wei Lee\textsuperscript{1}},
\\
 \textsuperscript{1}Singapore University of Technology and Design
\\
 \small{
   \{mingshan\_hee@mymail., aditi\_kumaresan@, roy\_lee@\}sutd.edu.sg
 }
}

\begin{document}
\maketitle
\begin{abstract}
The widespread presence of hate speech on the internet, including formats such as text-based tweets and vision-language memes, poses a significant challenge to digital platform safety. Recent research has developed detection models tailored to specific modalities; however, there is a notable gap in transferring detection capabilities across different formats. This study conducts extensive experiments using few-shot in-context learning with large language models to explore the transferability of hate speech detection between modalities. Our findings demonstrate that text-based hate speech examples can significantly enhance the classification accuracy of vision-language hate speech. Moreover, text-based demonstrations outperform vision-language demonstrations in few-shot learning settings. These results highlight the effectiveness of cross-modality knowledge transfer and offer valuable insights for improving hate speech detection systems\footnote{GitHub: \url{https://github.com/Social-AI-Studio/Bridging-Modalities}}.

\end{abstract}

\section{Introduction}

\paragraph{Motivation.}

Hate speech in the online space appears in various forms, including text-based tweets and vision-language memes. Recent hate speech studies have developed models targeting specific modalities \cite{cao2023pro, awal2021angrybert}.
However, these approaches are often optimized to within-distribution data and fail to address zero-shot out-of-distribution scenarios.




The emergence of vision-language hate speech, which comprises text and visual elements, presents two significant challenges. First, there is a scarcity of datasets, as this area has only recently gained lots of attention. Second, collecting and using such data is complicated by copyright issues and increasingly stringent regulations on social platforms. Consequently, the limited availability of vision-language data hampers performance in out-of-distribution cases. In contrast, the abundance and diversity of text-based data offer a potential source for cross-modality knowledge transfer \cite{hee2024recent}.

\paragraph{Research Objectives.}

This paper investigates whether text-based hate speech detection capabilities can be transferred to multimodal formats. By leveraging the richness of text-based data, we aim to enhance the detection of vision-language hate speech, addressing current research limitations and improving performance in low-resource settings.

\paragraph{Contributions.}

This study makes the following key contributions: (i) We conduct extensive experiments evaluating the transferability of text-based hate speech detection to vision-language formats using few-shot in-context learning with large language models. (ii) We demonstrate that text-based hate speech examples significantly improve the classification accuracy of vision-language hate speech. (iii) We show that text-based demonstrations in few-shot learning contexts outperform vision-language demonstrations, highlighting the potential for cross-modality knowledge transfer. These contributions address critical gaps in existing research and provide a foundation for developing robust hate speech detection systems.

\section{Research Questions}
As all forms of hate speech share one definition, this study investigates the usefulness of using hate speech from one form, such as text-based hate speech, to classify hate speech in another form, such as vision-language hate speech. Working towards this goal, we formulate two research questions to guide our investigation.

\paragraph{RQ1: Does the text hate speech support set help with vision-language hate speech?} 

Visual-language hate speech presents a distinct challenge compared to text-based hate speech, as malicious messages can hide within visual elements or interactions between modalities. It remains uncertain whether text-based hate speech can be useful for classifying visual-language hate speech. We investigate this uncertainty by performing few-shot in-context learning on large language models. This method allows the model to learn from text-based hate speech demonstration examples before classifying visual-language hate speech instances. 

\paragraph{RQ2: How does the text hate speech support set fare against the vision-language hate speech support set?}

Intuitively, using vision-language hate speech demonstrations should result in superior performance. However, the effectiveness of text-based hate speech demonstrations compared to vision-language hate speech demonstrations remains an open question. To investigate this gap, we conducted another round of few-shot in-context learning on large language models with a vision-language hate speech support set.

\begin{table}[t]
\centering
  \small
  \begin{tabular}{c|cc|cc}
    \hline
     & \multicolumn{2}{c|}{\textbf{Support}} & \multicolumn{2}{c}{\textbf{Test}}\\
    Dataset & \# H & \# Non-H & \# H & \# Non-H\\
    \hline\hline
    Latent Hatred & 8189 & 13,921 & - & -\\
    FHM-FG & 3,007 & 5,493 & 246 & 254 \\
    MAMI & - & - & 500 & 500 \\
    \hline
\end{tabular}
\caption{Statistical distributions of datasets, where "H" represents Hate and "Non-H" represents non-hate }
  \label{tab:dataset}
\end{table}

\begin{table*}[!ht]
  \small
  \centering
  \begin{tabular}{cccccccccc}
    \toprule
    & & & & \multicolumn{3}{c}{\textbf{FHM}} & \multicolumn{3}{c}{\textbf{MAMI}} \\
    \cmidrule(lr){5-7} \cmidrule(lr){8-10}
    \textbf{Model} & \textbf{\# Shots} & \textbf{Dem. Samp.} & \textbf{Matching} & \textbf{Acc.} & \textbf{F1} & \textbf{\# Invalids} & \textbf{Acc.} & \textbf{F1} & \textbf{\# Invalids} \\
    \midrule
    \multirow{16}{3.5em}{\centering Mistral-7B}  & 0-shot & - & - & 0.614 & 0.594 & 0 & 0.619 & 0.568  & 0 \\
    \cmidrule{2-10}
    & \multirow{5}{3.5em}{\centering 4-shots} & Random & - & 0.618 & 0.613 & 0 & 0.655 & 0.636 & 0 \\
    & & TF-IDF & Text. & 0.634 & 0.634 & 0 & 0.653 & 0.649 & 0 \\   
    & & TF-IDF & Rationale & 0.618 & 0.618 & 0 & 0.662 & 0.658 & 0 \\ 
    & & BM-25  & Text. & \underline{0.658} & \underline{0.657} & 0 & 0.665 & 0.662 & 0 \\    
    & & BM-25 & Rationale & \red{0.598} & \red{0.596} & 0 & \underline{0.676} & \underline{0.671} & 0 \\
    \cmidrule{2-10}
    & \multirow{5}{3.5em}{\centering 8-shots} & Random & - & 0.620 & 0.611 & 0 & 0.634 & 0.602 & 0 \\
    & & TF-IDF & Text. & 0.642 & 0.641 & 0 & 0.665 & 0.658 & 0 \\   
    & & TF-IDF & Rationale & 0.626 & 0.625 & 0 & 0.657 & 0.649 & 0 \\ 
    & & BM-25 & Text. & \textbf{\underline{0.660}} & \textbf{\underline{0.658}} & 0 & \underline{0.685} & \underline{0.680} & 0 \\    
    & & BM-25 & Rationale & \red{0.612} & 0.608 & 0 & 0.669 & 0.661 & 0 \\
    \cmidrule{2-10}
    & \multirow{5}{3.5em}{\centering 16-shots} & Random & - & 0.618 & 0.610 & 0 & 0.642 & 0.611 & 0 \\
    & & TF-IDF & Text. & \underline{0.644} & \underline{0.644} & 0 & 0.675 & 0.668 & 0 \\   
    & & TF-IDF & Rationale & 0.632 & 0.631 & 0 & 0.632 & 0.631 & 0\\ 
    & & BM-25 & Text. & 0.638 & 0.636 & 0 & \textbf{\underline{0.705}} & \textbf{\underline{0.701}} & 0 \\    
    & & BM-25 & Rationale & 0.614 & 0.611 & 0 & 0.665 & 0.659 & 0 \\  
    \midrule
    \multirow{16}{3.5em}{\centering Qwen2-7B}  & 0-shot & - & - & 0.624 & 0.609 & 0 & 0.614 & 0.574 & 0 \\
    \cmidrule{2-10}
    & \multirow{5}{3.5em}{\centering 4-shots} & Random & - & \red{0.620} & 0.614 & 0 & 0.653 & 0.632 & 0 \\ 
    & & TF-IDF & Text. & 0.632 & 0.631 & 0 & 0.650 & 0.641 & 0 \\   
    & & TF-IDF & Rationale & 0.634 & 0.633 & 0 & 0.663 & 0.653 & 0 \\ 
    & & BM-25 & Text. & \underline{0.644} & \underline{0.642} & 0 & \underline{0.672} & \underline{0.664} & 0 \\    
    & & BM-25 & Rationale & \red{0.590} & \red{0.587} & 0 & 0.663 & 0.654 & 0 \\
    \cmidrule{2-10}
    & \multirow{5}{3.5em}{\centering 8-shots} & Random & - & 0.632 & 0.628 & 0 & 0.645 & 0.622 & 0  \\ 
    & & TF-IDF & Text. & 0.632 & 0.632 & 0 & 0.656 & 0.650 & 0 \\   
    & & TF-IDF & Rationale & \red{0.618} & 0.617 & 0 & 0.664 & 0.656 & 0\\ 
    & & BM-25 & Text. & \textbf{\underline{0.654}} & \textbf{\underline{0.653}} & 0 & \textbf{\underline{0.679}} & \textbf{\underline{0.674}} & 0 \\    
    & & BM-25 & Rationale & \red{0.604} & 0.603 & 0 & 0.654 & 0.646 & 0 \\
    \cmidrule{2-10}
    & \multirow{5}{3.5em}{\centering 16-shots} & Random & - & 0.632 & 0.626 & 0 & 0.652 & 0.631 & 0 \\
        & & TF-IDF & Text. & 0.628 & 0.628 & 0 & 0.656 & 0.651 & 0 \\   
    & & TF-IDF & Rationale & \underline{0.632} & \underline{0.631} & 0 & 0.665 & 0.659 & 0 \\ 
    & & BM-25 & Text. & 0.624 & 0.624 & 0 & 0.678 & 0.674 & 0 \\    
    & & BM-25 & Rationale & 0.630 & 0.629 & 0 & \textbf{\underline{0.679}} & \textbf{\underline{0.674}} & 0 \\
    \bottomrule
    
  \end{tabular}
  \caption{Comparison of zero-shot and few-shot in-context learning with \textit{Latent Hatred} support set across different demonstration sampling (Dem. Sampl.) strategies. \underline{Underlined} represent the best results within a dataset for the given model and given few-shot setting, \textbf{bold} indicate the best results within a dataset for a given model across all few-shot settings and \textcolor{red}{red} denote few-shot in-context learning results below zero-shot performance.}
  \label{tab:lh-support-set-experiments}
\end{table*}

\begin{table*}[!ht]
  \small
  \centering
  \begin{tabular}{cccccccccc}
    \toprule
    & & & & \multicolumn{3}{c}{\textbf{FHM}} & \multicolumn{3}{c}{\textbf{MAMI}} \\
    \cmidrule(lr){5-7} \cmidrule(lr){8-10}
    \textbf{Model} & \textbf{\# Shots} & \textbf{Dem. Samp.} & \textbf{Matching} & \textbf{Acc.} & \textbf{F1} & \textbf{\# Invalids} & \textbf{Acc.} & \textbf{F1} & \textbf{\# Invalids} \\
    \midrule
    \multirow{16}{3.5em}{\centering Mistral-7B} & 0-shot & - & - & 0.614 & 0.594 & 0 & 0.619 & 0.568 & 0 \\
    \cmidrule{2-10}
    & \multirow{5}{3.5em}{\centering 4-shots} & Random & - & \underline{0.622} & \underline{0.617} & 0 & 0.656 & 0.642 & 0 \\
        & & TF-IDF & Text + Cap. & \red{0.604} & 0.598 & 0 & \underline{0.678} & \underline{0.670} & 0 \\   
    & & TF-IDF & Rationale & 0.618 & 0.613 & 0 & 0.662 & 0.652 & 0 \\ 
    & & BM-25 & Text + Cap. & \red{0.592} & \red{0.584} & 0 & 0.662 & 0.653 & 0 \\    
    & & BM-25 & Rationale & 0.620 & 0.617 & 0 & 0.667 & 0.659 & 0 \\
    \cmidrule{2-10}
    & \multirow{5}{3.5em}{\centering 8-shots} & Random & - & 0.624 & 0.615 & 0 & 0.652 & 0.632 & 0 \\
        & & TF-IDF & Text + Cap. & 0.618 & 0.611 & 0 & 0.675 & 0.664 & 0 \\   
    & & TF-IDF & Rationale & 0.628 & 0.622 & 0 & \textbf{\underline{0.681}} & \textbf{\underline{0.670}} & 0 \\ 
    & & BM-25 & Text + Cap. & \red{0.606} & 0.599 & 0 & 0.672 & 0.661 & 0 \\    
    & & BM-25 & Rationale & \underline{0.628} & \underline{0.624} & 0 & 0.674 & 0.666 & 0 \\
    \cmidrule{2-10}
    & \multirow{5}{3.5em}{\centering 16-shots} & Random & - & 0.620 & 0.614 & 0 & 0.668 & 0.651 & 0 \\
        & & TF-IDF & Text + Cap. & 0.620 & 0.617 & 0 & 0.672 & 0.665 & 0 \\   
    & & TF-IDF & Rationale & \textbf{\underline{0.638}} & \textbf{\underline{0.635}} & 0 & 0.671 & 0.661 & 0 \\ 
    & & BM-25 & Text + Cap. & 0.630 & 0.625 & 0 & 0.682 & 0.673 & 0 \\    
    & & BM-25 & Rationale & 0.634 & 0.633 & 0 & \underline{0.687} & \underline{0.680} & 0 \\
    \midrule
    \multirow{16}{3.5em}{\centering Qwen2-7B}  & 0-shot & - & - & 0.624 & 0.609 & 0 & 0.614 & 0.574 & 0 \\
    \cmidrule{2-10}
    & \multirow{5}{3.5em}{\centering 4-shots} & Random & - & \red{0.606} & \red{0.602} & 0 & 0.655 & 0.642 & 0 \\
        & & TF-IDF & Text + Cap. & \red{0.620} & 0.620 & 0 & 0.659 & 0.657 & 0 \\   
    & & TF-IDF & Rationale & \textbf{\underline{0.636}} & \textbf{\underline{0.636}} & 0 & 0.650 & 0.646 & 0 \\ 
    & & BM-25 & Text + Cap. & \red{0.616} & 0.616 & 0 & \textbf{\underline{0.676}} & \textbf{\underline{0.674}} & 0 \\    
    & & BM-25 & Rationale & \red{0.622} & 0.622 & 0 & 0.669 & 0.672 & 0 \\
    \cmidrule{2-10}
    & \multirow{5}{3.5em}{\centering 8-shots} & Random & - & \red{0.592} & \red{0.581} & 0 & 0.642 & 0.624 & 0 \\
        & & TF-IDF & Text + Cap. & \red{0.606} & \red{0.604} & 0 & 0.648 & 0.645 & 0 \\   
    & & TF-IDF & Rationale & \red{0.620} & 0.619 & 0 & 0.649 & 0.644 & 0 \\ 
    & & BM-25 & Text + Cap. & \red{0.614} & 0.613 & 0 & 0.665 & 0.662 & 0 \\    
    & & BM-25 & Rationale & \underline{0.624} & \underline{0.623} & 0 & \underline{0.669} & \underline{0.664} & 0 \\
    \cmidrule{2-10}
    & \multirow{5}{3.5em}{\centering 16-shots} & Random & - & \red{0.602} & \red{0.592} & 0 & 0.650 & 0.634 & 0 \\
        & & TF-IDF & Text + Cap. & \red{0.610} & 0.610 & 0 & 0.649 & 0.648 & 0 \\   
    & & TF-IDF & Rationale & \red{0.604} & \red{0.604} & 0 & \underline{0.656} & \underline{0.653} & 0 \\ 
    & & BM-25 & Text + Cap. & \red{0.610} & 0.610 & 0 & 0.654 & 0.653 & 0 \\    
    & & BM-25 & Rationale & \underline{0.626} & \underline{0.626} & 0 & 0.653 & 0.650 & 0 \\
    \bottomrule
  \end{tabular}
  \caption{Comparison of zero-shot and few-shot in-context learning experiment results with \textit{FHM} support set across different demonstration sampling (Dem. Sampl.) strategies. \underline{Underlined} represent the best results within a dataset for the given model and given few-shot setting, \textbf{bold} indicate the best results within a dataset for a given model across all few-shot settings and \textcolor{red}{red} denote few-shot in-context learning results below zero-shot performance.}
  \label{tab:fhm-support-set-experiments}
\end{table*} 

\section{Experiments}

\subsection{Experiment Settings}

\paragraph{Models.} 

We use the Mistral-7B\footnote{mistralai/Mistral-7B-Instruct-v0.3} \cite{jiang2023mistral} and Qwen2-7B\footnote{Qwen/Qwen2-7B-Instruct} \cite{qwen} models, both of which demonstrate strong performance across various benchmarks, in our primary experiments. Notably, their models on LMSYS's Chatbot Arena Leaderboard achieve high ELO scores \cite{chiang2024chatbot}. To facilitate reproducibility and minimize randomness, we use the greedy decoding strategy for text generation.

We conducted additional experiments to support the findings in our paper further with two additional models: LLaVA-7B\footnote{llava-hf/llava-v1.6-mistral-7b-hf} and Llama3-8B\footnote{meta-llama/Llama-3.1-8B-Instruct}. The results of these experiments are presented in Appendix \ref{appendix:additional-experiments}.

\paragraph{Test Datasets.}
The Facebook Hateful Memes (FHM) dataset \cite{mathias2021fhmfg} contains synthetic memes categorized into five types of hate incitement: gender, racial, religious, nationality, and disability-based.  The Multimedia Automatic Misogyny Identification (MAMI) \cite{fersini2022mami} dataset comprises real-world misogynistic memes classified into shaming, stereotype, objectification, and violence categories. Both datasets contain text overlay information, eliminating the need for an OCR model to extract text.

For evaluation, we use the FHM's \textit{dev\_seen} split, which includes 246 hateful memes and 254 non-hateful ones, and the MAMI's \textit{test} split, consisting of 500 hateful and 500 non-hateful memes.


\paragraph{Text Support Set.}
We use the Latent Hatred \cite{elsherief2021latent} dataset, which includes both explicit and implicit forms of hate speech, such as coded and indirect derogatory attacks. This dataset comprises 13,921 non-hateful speeches, 1,089 explicit hate speeches, and 7,100 implicit hate speeches. 

\paragraph{Vision-Language Support Set.}
We use the FHM train split for evaluation, containing 3,007 hateful memes and 5,493 non-hateful memes.

\subsection{Data Preprocessing}

\paragraph{Image Captioning.}
To perform hateful meme classification with the large language models, we perform image captioning on the meme using the OFA \cite{wang2022ofa} model pre-trained on the MSCOCO \cite{lin2014microsoft} dataset. 

\paragraph{Rationale Generation.}

We prompt Mistral-7B to generate informative rationales that explain the underlying meaning of the content, providing additional context for the few-shot in-context learning. Specifically, the model generates rationales by using the content and ground truth labels (i.e., prompt + content $\rightarrow$ ground truth label $\rightarrow$ explanation). For the Latent Hatred dataset, we use post information and labels, while for the FHM dataset, we use meme text, captions, and labels. To mitigate noise from varying rationale formulations, we instruct the model to consider both textual and visual elements, focusing on target groups, imagery, and the impact of tweet/meme bias perpetuation. More details can be found in Appendix \ref{sec:appendix-rationale-generation}.

\subsection{RQ1: Does text hate speech help with vision-language hate speech?}
To evaluate the effectiveness of the few-shot in-context learning approach and the Latent Hatred support set, we employed three sampling strategies: Random sampling, TF-IDF sampling, and BM-25 sampling. The TF-IDF and BM-25 strategies leverage the text and caption information of the test record to identify similar examples from the support set, focusing on either the text or the generated rationale. Table \ref{tab:lh-support-set-experiments} shows the comparison of zero-shot and few-shot in-context learning experiment results with Latent Hatred support set.

The experimental results demonstrate that employing a few-shot in-context learning approach with text-based hate speech demonstrations is highly effective in classifying vision-language hate speech. Firstly, while the random sampling strategy could retrieve more irrelevant demonstrations compared to other strategies, the few-shot in-context learning with random sampling surpasses the zero-shot inference performance on both models across two datasets in terms of F1 score. Secondly, the TF-IDF and BM-25 sampling strategies exceed the zero-shot inference performance on both models within the MAMI dataset. Conversely, within the FHM dataset, we observed several instances where some sampling strategies in the few-shot in-context learning scenario performed worse than zero-shot inference. However, these sampling strategies consistently outperformed zero-shot inference when run with 16-shots in-context learning. Lastly, the best few-shot in-context learning performance within each dataset and each model shows significant improvement over zero-shot model performance. For example, the Mistral-7B model achieves an F1 score improvement of 0.64 and 1.23 on the FHM and MAMI datasets respectively.

\subsection{RQ2: How does text hate speech support set fare against vision-language hate speech support set?}
 
 Table \ref{tab:fhm-support-set-experiments} shows the comparison of zero-shot and few-shot in-context learning experiment results with the FHM support set. The experimental results indicate that using the FHM support set can enhance model performance in some scenarios. However, it is noteworthy that in many instances, few-shot in-context learning performs worse than zero-shot model performance when compared against the Latent Hatred support set. Most significantly, the model encounters the most failures on the FHM test set despite using the FHM train set as a support set. We also observed that the best model performance with the Latent Hatred support set surpasses the best model performance with the FHM support set across all instances. We speculate that this discrepancy may stem from the oversimplification of visual information into image captions and the broader topic coverage provided by the Latent Hatred dataset. Nevertheless, this suggests that text-based data can serve as a valuable resource for improving performance on multimodal tasks, particularly in low-resource settings.

\section{Few-Shot Demonstration Analysis}

While including relevant few-shot in-context learning examples can improve model performance, the degree to which these examples benefit the model remains uncertain. To gain deeper insights, we examine the examples that got correctly classified and misclassified using the demonstration exemplars from the Latent Hatred support dataset.


The detailed analysis and case study examples, along with their few-shot in-context demonstrations, can be found in Appendix \ref{appendix:demonstration-analysis} and \ref{appendix:in-context-demonstrations}.

\paragraph{Latent Hatred's Support Set}
We found that using relevant examples as demonstrations significantly improves classification, as the additional context aids the model in evaluating similar content more effectively. This approach enhances the model’s ability to generalize across diverse hate speech contexts and formats, thereby helping to reduce false negatives in edge cases. However, we also observed that models sometimes misinterpret neutral content as hateful. This misinterpretation may arise from exposure to demonstration examples that contain dismissive or derogatory language on sensitive topics. Consequently, these examples can lead to an overgeneralization of what qualifies as hateful, causing content that was correctly classified in a zero-shot setting to be misclassified. This issue is similar to the problem of oversensitivity to specific terms found in fine-tuned multimodal hate speech detection models \cite{cuo2022understanding, hee2022explaining, rizzi2023recognizing}.

\section{Related Works}

Numerous approaches have been proposed to tackle the online hate speech problem \cite{cao2023pro, lee2021disentangling, hee2023decoding, lin2024towards}. While these approaches demonstrate impressive performance, they often require large amounts of data for fine-tuning, and the rapid evolution of hate speech can quickly render these models outdated. Furthermore, a recent study indicated that these models are vulnerable to adversarial attacks \cite{aggarwal2023hateproof}.

These challenges led to exploring few-shot hate speech detection approaches, where models learn using limited data \cite{facebook2021fewshotlearner, awal2023model}. Mod-HATE trains specialized modules on related tasks and integrates the weighted module with large language models to enhance detection capabilities \cite{cao2024modularized}. Our approach contributes to this field by addressing the challenge of limited data availability, using the abundance and diversity of text-based hate speech as an alternative source for cross-modality knowledge transfer.

\section{Conclusion}
We investigated the possibility of cross-modality knowledge transfer using few-shot in-context learning with large language models. Our extensive experiments show that text-based hate speech demonstrations significantly improve the classification accuracy of vision-language hate speech, and using text-based demonstrations in few-shot in-context learning outperforms using vision-language demonstrations. For future works, we aim to extend our analysis to more datasets and explore other cross-modality knowledge transfer approaches such as cross-modality fine-tuning.


\section*{Acknowledgement}
This research is supported by the Ministry of Education, Singapore, under its Academic Research Fund Tier 2 (Award ID: MOE-T2EP20222-0010). Any opinions, findings and conclusions or recommendations expressed in this material are those of the authors and do not reflect
the views of the Ministry of Education, Singapore.

\section*{Limitations}

There are several limitations in this research study

\textbf{Model Coverage and Model Size.} In this study, we evaluated and compared two large models containing 7B parameters. In the future, we aim to extend our analysis to other large models when more computational resources are available. 

\textbf{Large Language Model.} In this study, we evaluated few-shot in-context learning in large language models. The experiments are designed in this manner, so to ensure that there can be a fair comparison between the different support sets. We recognize that using a vision-language support set for few-shot in-context learning with a large vision-language model could achieve better performance. However, evaluation using large vision-language models would then be unfair to text support set for few-shot in-context learning.

\section*{Ethical Considerations}

\paragraph{Impact of False Positives.} Developing a reliable and generalizable hate speech detection system is crucial, as false positives can significantly impact free speech and diminish user trust. Firstly, overly aggressive detection systems may mistakenly flag content that does not qualify as hate speech, thereby suppressing free speech and hindering meaningful discussions. Secondly, when users frequently encounter false positives, their confidence in the platform's moderation system may diminish. The reduced trust can result in decreased user engagement and a perception of bias within the platform.

\bibliography{custom}

\appendix

\section{Potential Risks}
This project seeks to counteract the dissemination of harmful memes, aiming to protect individuals from prejudice and discrimination based on race, religion and gender. However, we acknowledge the risk of malicious users reverse-engineering memes to evade detection by CMTL-RAG AI systems, which is strongly discouraged and condemned.

\section{Licenses and Usage Scientific Artifacts}

\subsection{Models}

All of the LLMs used in this paper contain licenses permissive for academic and/or research use.

\begin{itemize}
    \item \textbf{Mistral-7B} Apache-2.0 License
    \item \textbf{Qwen2-7B} Apache-2.0 License
    \item \textbf{LLaVA-7B} Apache 2.0 License 
    \item \textbf{LLaMA-8B} Llama 3.1 Community License
\end{itemize}

\subsection{Datasets}

All of the datasets used in this paper contain licenses permissive for academic and/or research use.

\begin{itemize}
    \item \textbf{Latent Hatred Dataset.} MIT License
    \item \textbf{Hateful Memes Dataset.} MIT License
    \item \textbf{Multimedia Automatic Misogyny Identification.} Creative Commons License (CC BY-NC-SA 4.0)
\end{itemize}
\subsection{Anonymity and Offensive Content}
The datasets used in this research contain offensive content, which is crucial for addressing the research questions. Importantly, there are no unique identifiers for the individuals who authored the hateful content in these datasets.


\section{Computational Experiments}
NVIDIA A40 GPUs were utilized for the work done in this paper.

\subsection{Experimental Setup}
We thoroughly discussed the experimental setup in the main body of the paper. This included descriptions of the models used (Mistral-7B and Gwen2), the number of shots (0-shot, 4-shots, 8-shots, 16-shots), and the different strategies employed for matching (Random, TF-IDF, BM-25) across two datasets (FHM and MAMI). Best-found hyperparameter values were highlighted in the results tables, such as the highest accuracy and F1 scores achieved for each experimental condition.

\subsection{Use of Existing Packages}
\subsubsection{Large Language Models}
\begin{itemize}
    \item \textbf{transformers 4.41.1}
\end{itemize}
\subsubsection{Matching and Retrieval Scoring}
\begin{itemize}
    \item \textbf{rank-bm25 0.2.2} for BM-25 similarity matching based on this implementation \href{https://pypi.org/project/rank-bm25/}{link}.
    \item \textbf{scikit-learn 1.5.0} for TF-IDF similarity matching based on this implementation \href{https://scikit-learn.org/stable/whats_new/v1.5.html#}{link} .
\end{itemize}

\section{Few-Shot In-Context Learning}

In this approach, we retrieve relevant labelled examples from a 'support dataset' using similarity metrics such as TF-IDF or BM-25 for a given meme from the inference dataset. These examples are then provided as demonstrations in a few-shot prompt to enhance the model's understanding of the meme. Finally, we prompt the model to classify the meme, leveraging the augmented context for improved accuracy.

\section{Similarity Metrics}
\subsection{TF-IDF}
TFIDF is a statistical measure used to evaluate the importance of a word in a document relative to a collection of documents (corpus). By creating TF-IDF vectors for a 'support dataset', we can use cosine similarity to find the most similar records to a given inference record.
\subsection{BM25}
BM25 is an advanced version of the TF-IDF weighting scheme used in search engines. It incorporates term frequency saturation and document length normalization to improve retrieval performance. We generate vectors for each record in the 'support dataset' and use cosine similarity to identify records most similar to an inference record. 

\section{Rationale Generation Details}
\label{sec:appendix-rationale-generation}
We use the Mistral-7B model, a state-of-the-art language model known for its capabilities in language understanding and generation. We implement a ten-shot prompting method to generate explanations for the hateful tweets in the Latent Hatred dataset. Specifically, we select five examples of hateful posts and five non-hateful posts for the ten-shot prompt demonstrations. Each demonstration in the prompt follows the following template, given the post text, the post label (hateful or not hateful) and the post rationale: \newline\newline
\textbf{User:} \textit{Determine whether the following post is hateful. Text: }\{text\}:  \newline
\textbf{Assistant:} \{label\} \newline
\textbf{User:} \textit{Briefly provide an explanation, in no more than three points, for the post being perceived as \{label\}. Your explanation should address the targeted group, any derogatory imagery or language used, and the impact it has on perpetuating bias, stereotypes, prejudice, discrimination or inciting harm.} \newline
\textbf{Assistant:} \textit{Answer:} \{rationale\} \newline\newline
Similarly, for FHM/MAMI, we select five examples of hateful memes and five non-hateful memes for the ten-shot prompt demonstrations. Each demonstration in the prompt follows the following template, given the meme text, the meme label (hateful or not hateful) and rationale: \newline\newline
\textbf{User:} \textit{Determine whether the following meme is hateful. Text: }\{text\} \textit{Caption: } \{caption\} \newline
\textbf{Assistant:} \{label\} \newline
\textbf{User:} \textit{Briefly provide an explanation, in no more than three points, for the meme being perceived as \{label\}. Your explanation should address the targeted group, any derogatory imagery or language used, and the impact it has on perpetuating bias, stereotypes, prejudice, discrimination or inciting harm.} \newline
\textbf{Assistant:} \textit{Answer:} \{rationale\} \newline \newline
The demonstration explanation follows a list format, where each list item addresses the targeted group, any derogatory imagery or language used, and the impact it has on perpetuating bias. Finally, to reiterate the classification of the post's hatefulness, the explanation concludes with the sentence, \textit{"In summary, this post/meme is} \{label\}". During inference, the demonstrations provide a structured guide for the model to generate rationales based on the given post/meme's text and label.

\section{In-Context Demonstration Analysis}
\label{appendix:demonstration-analysis}
This section evaluates and compares the effectiveness of Latent Hatred in enhancing the models' multimodal hateful memes classification ability. Additionally, we identify situations where the context provided by these demonstrations can sometimes hinder the model's performance. The in-context demonstrations for each case study example can be found in Appendix \ref{appendix:in-context-demonstrations}.

\subsection{Case Study: Latent Hatred Cross-Modality Effectiveness}
\paragraph{Latent Hatred - Correct Classifications.}We examined and studied two cases where the Mistral-7B model failed to correctly classify the FHM meme in the 0-shot classification setting but succeeded when latent hatred demonstrations were introduced at 4, 8, and 16-shot levels. 

\begin{table}[!ht]
\centering
\small
\begin{tabular}{p{0.90\linewidth}}
\toprule
\textbf{Meme Text:} life hack \#23 how to get stoned with no weed\\
\textbf{Meme Caption:} a young woman in a hijab kisses her mother on the cheek\\
\textbf{Label:} Hateful\\
\bottomrule
\end{tabular}
\caption{Correct Classification - Case Study 1}
\label{fig:cs1-p1-example1}
\end{table}

\paragraph{Example 1 - Analysis.} Demonstration 1 is crucial because it demonstrates how derogatory language targeting a religious group (comparing the Qur'an to marijuana) directly relates to the meme's use of sensitive imagery (a woman in a hijab) and the potential for offending religious or cultural groups.

\begin{table}[!ht]
\centering
\small
\begin{tabular}{p{0.90\linewidth}}
\toprule
\textbf{Meme Text:} when you have an 80 point iq but it's the highest in your village\\
\textbf{Meme Caption:} a photograph of a woman with an afro\\
\textbf{Label:} Hateful\\
\bottomrule
\end{tabular}
\caption{Correct Classification - Case Study 2}
\label{fig:cs1-p1-example2}
\end{table}

\paragraph{Example 2 - Analysis.} The meme implies that people with an afro, ‘presumably from a poor country’, have low IQs and that an 80 IQ is the best they can achieve due to poverty. Demonstration 2 highlights how generalizations about IQ can perpetuate stereotypes about intelligence and economic status. 

\paragraph{Latent Hatred - Incorrect Classifications.3} We examined and studied two cases where the model's initial success in 0-shot classification declined when Latent Hatred demonstrations were included in 4, 8, and 16-shot settings. 

\begin{table}[!ht]
\centering
\small
\begin{tabular}{p{0.90\linewidth}}
\toprule
\textbf{Meme Text:} meanwhile in baltimore\\
\textbf{Meme Caption:} a baboon mounting another baboon in the serengeti\\
\textbf{Label:} Not Hateful\\
\bottomrule
\end{tabular}
\caption{Incorrect Classification - Case Study 1}
\label{fig:cs1-p2-example1}
\end{table}

\paragraph{Example 2 Analysis.} Demonstration 1 features derogatory language that dehumanizes individuals by comparing them to baboons, which likely influenced the model to associate the meme's content with hatefulness. The model might have incorrectly applied the same reasoning to the meme with the baboon imagery, interpreting it as offensive or derogatory. 

\begin{table}[!ht]
\centering
\small
\begin{tabular}{p{0.90\linewidth}}
\toprule
\textbf{Meme Text:} you can smell disgust and hate from this photo\\
\textbf{Meme Caption:} a black and white photo of two men standing next to each other\\
\textbf{Label:} Not hateful\\
\bottomrule
\end{tabular}
\caption{Incorrect Classification - Case Study 2}
\label{fig:cs1-p2-example2}
\end{table}

\paragraph{Analysis.}  The model misclassified the meme as hateful due to its reliance on examples like Example 3 and Example 4, which have similar wording, leading to a misinterpretation of the neutral photo. Example 3 involves dismissive language that trivializes hate crimes, potentially influencing the model to perceive similar tones or implications in the meme. Example 4 demonstrates how derogatory comparisons involving sensitive topics (like Jewish people and the Holocaust) can be classified as hateful.

\onecolumn
\section{In-context Demonstrations}
\label{appendix:in-context-demonstrations}

\subsection{Latent Hatred Support Set - Correct Classification Case Study \#1}
\begin{table*}[h]
\small
\centering
\begin{tabular}{p{0.95\linewidth}}
\toprule
\textbf{\#\#\# EXAMPLE 1}\\
\textbf{Content}: A post containing 'my weed is like the qur'an . <repeat> i burn it until i get stoned .'\\
\textbf{Answer}: Hateful\\
\textbf{Rationale}: Targeted Group: The post targets individuals who follow Islam by making a derogatory comparison between their holy book, the Qur'an, and marijuana.\\
Derogatory Imagery/Language: The statement uses disrespectful and offensive language by comparing the Qur'an to a drug and suggesting that it should be burned.\\
Impact on Bias/Stereotypes: This post reinforces harmful stereotypes and prejudices against Muslims by belittling their religious beliefs and promoting disrespect towards their holy book. It could incite fear and hostility towards Muslim individuals and communities.\\
In summary, this post is hateful.\\ \\

\textbf{\#\#\# EXAMPLE 2}\\
\textbf{Content}: A post containing ' " vile weed ! " " " '\\
\textbf{Answer}: Not Hateful\\
\textbf{Rationale}: Targeted Group: The post does not target a specific group based on attributes such as race, religion, ethnicity, gender, sexual orientation, disability, or other distinguishing characteristics.\\
Derogatory Imagery/Language: There is no derogatory language or imagery used in the post. The phrase "vile weed" could be interpreted as negative, but it does not specifically target any group.\\
Impact on Bias/Stereotypes: The post does not perpetuate harmful stereotypes or prejudices against any group. It does not contribute to bias, discrimination, or inciting harm.\\
In summary, this post is not hateful.\\ \\

\textbf{\#\#\# EXAMPLE 3}\\
\textbf{Content}: A post containing '> > > americas future > > > white woman raped then stoned by racist blacks in south africa'\\
\textbf{Answer}: Hateful\\
\textbf{Rationale}: Targeted Group: The post targets black individuals, particularly those in South Africa, by implying that they are responsible for a violent crime against a white woman.\\
Derogatory Imagery/Language: The use of the phrase "racist blacks" is derogatory and perpetuates harmful stereotypes about black people, suggesting that they are inherently violent and discriminatory. The title "america's future" also implies that the safety and well-being of white people is at risk due to the actions of black individuals.\\
Impact on Bias/Stereotypes: This post reinforces harmful stereotypes and prejudices against black individuals, potentially inciting fear and hostility towards them. It contributes to a narrative that portrays black people as dangerous and threatening, which can perpetuate racism and discrimination.\\
In summary, this post is hateful.\\ \\

\textbf{\#\#\# EXAMPLE 4}\\
\textbf{Content}: A post containing 'mandela's rainbow nation : white woman raped then stoned by racist blacks in south africa via'\\
\textbf{Answer}: Hateful\\
\textbf{Rationale}: Targeted Group: The post targets black individuals in South Africa, implying that they are violent and racist towards white women.\\
Derogatory Imagery/Language: The statement uses derogatory language by describing black individuals as "racist" and suggesting that they are responsible for a violent act against a white woman.\\
Impact on Bias/Stereotypes: This post reinforces harmful stereotypes and prejudices against black individuals, perpetuating the idea that they are violent and dangerous towards white people. It could incite fear and hostility towards black South Africans while promoting discriminatory and divisive beliefs.\\
In summary, this post is hateful.\\ \\

\bottomrule
\end{tabular}
\label{tab:cs1-p1-dem1}
\end{table*}

\newpage
\subsection{Latent Hatred Support Set - Correct Classification Case Study \#2}
\begin{table*}[h]
\small
\centering
\begin{tabular}{p{0.95\linewidth}}
\toprule
\textbf{\#\#\# EXAMPLE 1}\\
\textbf{Content:} A post containing 'yup . anyone with an iq above 80 can figure out what happened .' \\
\textbf{Answer:} Not Hateful\\
\textbf{Rationale:} Targeted Group: The post does not target a specific group based on attributes such as race, religion, ethnicity, gender, sexual orientation, disability, or other distinguishing characteristics.\\

Derogatory Imagery/Language: There is no derogatory language or imagery used in the post.\\

Impact on Bias/Stereotypes: The post does not perpetuate harmful stereotypes or prejudices against any group. Instead, it seems to be making a general comment about intelligence. However, it is important to note that using the term "IQ" can be perceived as insensitive, as it may be associated with harmful stereotypes about certain racial or ethnic groups.\\

In summary, this post is not hateful, but it is important to be mindful of the potential implications of using terms like "IQ" in a way that could be perceived as insensitive or reinforcing harmful stereotypes.\\ \\

\textbf{\#\#\# EXAMPLE 2}\\
\textbf{Content:} A post containing 'believe it or not  iq matters . it's no suprise the highest iq nations are the most prosperous and vice versa .' \\
\textbf{Answer:} Not Hateful \\
\textbf{Rationale:} Targeted Group: The post does not target a specific racial, ethnic, or other group with harmful intent. \\

Derogatory Imagery/Language: There is no derogatory language or imagery used in the post. \\

Impact on Bias/Stereotypes: The post discusses IQ and its relationship to prosperity, but it does not perpetuate harmful stereotypes or prejudices against any group. It does not promote discrimination or incite harm. Instead, it presents a general observation about the relationship between IQ and prosperity. However, it is essential to note that intelligence is a complex trait influenced by various factors, and it is crucial to avoid using it to stereotype or discriminate against individuals or groups. \\

In summary, this post is not hateful. \\ \\

\textbf{\#\#\# EXAMPLE 3}\\
\textbf{Content:} A post containing 'france 80 \% white ? ?' \\
\textbf{Answer:} Not Hateful \\
\textbf{Rationale:} Targeted Group: The post does not explicitly target a specific racial or ethnic group with harmful intent. \\

Derogatory Imagery/Language: There is no derogatory language or imagery used in the post. \\

Impact on Bias/Stereotypes: The post does not perpetuate harmful stereotypes or prejudices against any group. It appears to be a question about demographic statistics, not an attempt to promote bias or discrimination. \\

In summary, this post is not hateful. \\ \\

\textbf{\#\#\# EXAMPLE 4}\\
\textbf{Content:} A post containing 'sorry muslim not the highest' \\
\textbf{Answer:} Hateful \\
\textbf{Rationale:} Targeted Group: The post targets Muslims by implying that they are inferior or less valued. \\

Derogatory Imagery/Language: The statement uses a derogatory tone towards Muslims by suggesting that they are not the "highest." \\

Impact on Bias/Stereotypes: This statement reinforces harmful stereotypes and prejudices against Muslims, potentially inciting fear and hostility towards them. It contributes to an environment that doesn't respect or value the beliefs and practices of this religious group. \\

In summary, this post is hateful. \\ \\

\bottomrule
\end{tabular}
\label{tab:cs1-p1-dem2}
\end{table*}

\newpage
\subsection{Latent Hatred Support Set - Incorrect Classification Case Study \#1}
\begin{table*}[h]
\small
\centering
\begin{tabular}{p{0.95\linewidth}}
\toprule
\textbf{\#\#\# EXAMPLE 1}\\
\textbf{Content:} A post containing 'you no doubt  are a baboon  in appearance and in iq .'\\
\textbf{Answer:} Hateful\\
\textbf{Rationale:} Targeted Group: The post targets individuals with a lower intelligence quotient (IQ), which can be associated with various groups, including people with intellectual disabilities, racial or ethnic minorities, or individuals from lower socioeconomic backgrounds.\\

Derogatory Imagery/Language: The statement uses derogatory language by comparing the person to a baboon, which is a species often stereotyped as less intelligent. The use of "baboon" is dehumanizing and disrespectful.\\

Impact on Bias/Stereotypes: This post reinforces harmful stereotypes and prejudices by associating lower intelligence with a specific group of people. It contributes to an environment that doesn't respect or support individuals with lower IQs and may incite discrimination or harm against them.\\

In summary, this post is hateful.\\ \\

\textbf{\#\#\# EXAMPLE 2}\\
\textbf{Content:} A post containing 'yes ! meanwhile in the anti-white'\\
\textbf{Answer:} Not Hateful\\
\textbf{Rationale:} Targeted Group: The post does not explicitly target a specific racial or ethnic group with harmful intent.\\

Derogatory Imagery/Language: There is no derogatory language or imagery used in the post.\\

Impact on Bias/Stereotypes: The post does not perpetuate harmful stereotypes or prejudices against any group. Instead, it appears to be a response to a specific context or movement, without promoting discrimination or inciting harm.\\

In summary, this post is not hateful.\\ \\

\textbf{\#\#\# EXAMPLE 3}\\
\textbf{Content:} A post containing 'another muslim in the white house'\\
\textbf{Answer:} Hateful\\
\textbf{Rationale:} Targeted Group: The post targets Muslim individuals by implying that their religious identity is a threat to the "white house" and the broader society.\\

Derogatory Imagery/Language: The statement uses derogatory language by associating the Muslim community with a perceived threat to the "white house," which could be interpreted as a symbol of the United States or its political system.\\

Impact on Bias/Stereotypes: This post reinforces harmful stereotypes and prejudices against Muslim individuals, potentially inciting fear and hostility towards them. It contributes to an environment that doesn't respect or support the rights and beliefs of this marginalized group.\\

In summary, this post is hateful.\\ \\

\textbf{\#\#\# EXAMPLE 4}\\
\textbf{Content:} A post containing 'meanwhile  \#truconservatives in the gop are dropping opposition to gay marriage :'\\
\textbf{Answer:} Not Hateful\\
\textbf{Rationale:} Targeted Group: The post does not target a specific group with harmful intent; instead, it discusses a political issue related to the GOP and gay marriage.\\

Derogatory Imagery/Language: There is no derogatory language or imagery used in the post.\\

Impact on Bias/Stereotypes: The post does not perpetuate harmful stereotypes or prejudices against any group. It simply discusses a political issue and does not incite discrimination or harm.\\

In summary, this post is not hateful.\\ \\

\bottomrule
\end{tabular}
\label{tab:cs1-p2-dem1}
\end{table*}

\newpage
\subsection{Latent Hatred Support Set - Incorrect Classification Case Study \#2}
\begin{table*}[h]
\small
\centering
\begin{tabular}{p{0.95\linewidth}}
\toprule
\textbf{\#\#\# EXAMPLE 1}\\
\textbf{Content:} A post containing 'those people in the black and white photo were ahead of their time .'\\
\textbf{Answer:} Not Hateful\\
\textbf{Rationale:} Targeted Group: The post does not target a specific racial or ethnic group with harmful intent; rather, it comments on a historical photograph.\\

Derogatory Imagery/Language: There is no derogatory language or imagery used in the post.\\

Impact on Bias/Stereotypes: The post does not perpetuate harmful stereotypes or prejudices against any group. Instead, it highlights the progress made by individuals in the past, regardless of their race or ethnicity.\\

In summary, this post is not hateful. \\ \\

\textbf{\#\#\# EXAMPLE 2}\\
\textbf{Content:} A post containing 'the white resister ( ) posted a photo on twitter'\\
\textbf{Answer:} Not Hateful\\
\textbf{Rationale:} Targeted Group: The post does not explicitly target a specific group based on attributes such as race, religion, ethnicity, gender, sexual orientation, disability, or other distinguishing characteristics.\\

Derogatory Imagery/Language: There is no derogatory language or imagery used in the post.\\

Impact on Bias/Stereotypes: The post does not perpetuate harmful stereotypes or prejudices against any group. It simply mentions a Twitter user named "the white resister," which does not imply any negative intent or bias.\\

In summary, this post is not hateful.\\ \\

\textbf{\#\#\# EXAMPLE 3}\\
\textbf{Content:} A post containing '-sorry. this is a hoax photo. but aww, you want to believe it, don't you?'\\
\textbf{Answer:} Hateful\\
\textbf{Rationale:} Targeted Group: The post does not explicitly target a specific group, but it is likely to be perceived as offensive towards individuals who are victims of hate crimes or discrimination based on their race, religion, ethnicity, gender, sexual orientation, or other distinguishing characteristics.\\

Derogatory Imagery/Language: The statement "aww, you want to believe it, don't you?" is dismissive and potentially mocking, which could be interpreted as a form of derogatory language.\\

Impact on Bias/Stereotypes: The post may contribute to an environment that trivializes hate crimes and discrimination, potentially minimizing the experiences and struggles of marginalized groups. This could perpetuate bias, stereotypes, and prejudice against these groups.\\

In summary, the post is hateful as it may contribute to a dismissive and disrespectful attitude towards victims of hate crimes and discrimination.\\ \\

\textbf{\#\#\# EXAMPLE 4}\\
\textbf{Content:} A post containing 'What do you call a picture of smoke? A Jewish family photo'\\
\textbf{Answer:} Hateful\\
\textbf{Rationale:} Targeted Group: The post targets Jewish people by making a derogatory comparison between a Jewish family and smoke, which is a harmful and offensive stereotype.\\

Derogatory Imagery/Language: The statement uses a derogatory comparison that associates Jewish people with smoke, a symbol often used to represent the Holocaust and the persecution of Jewish people.\\

Impact on Bias/Stereotypes: This post reinforces harmful stereotypes and prejudices against Jewish people, potentially inciting fear and hostility towards them. It also contributes to a hostile and exclusionary environment for this marginalized group.\\

In summary, this post is hateful.\\ \\

\bottomrule
\end{tabular}
\label{tab:cs1-p2-dem2}
\end{table*}



\onecolumn
\section{Additional Experiments}
\label{appendix:additional-experiments}

\subsection{LLaVA-7B}
\label{appendix:llava-experiments}

\begin{table*}[h!]
  \small
  \centering
  \begin{tabular}{cccccccccc}
    \toprule
    & & & & \multicolumn{3}{c}{\textbf{FHM}} & \multicolumn{3}{c}{\textbf{MAMI}} \\
    \cmidrule(lr){5-7} \cmidrule(lr){8-10}
    \textbf{Model} & \textbf{\# Shots} & \textbf{Dem. Samp.} & \textbf{Matching} & \textbf{Acc.} & \textbf{F1} & \textbf{\# Invalids} & \textbf{Acc.} & \textbf{F1} & \textbf{\# Invalids} \\
    \midrule
    \multirow{16}{3.5em}{\centering LLaVA-7B} & 0-shot & - & - & 0.512 & 0.509 & 27 & 0.553 & 0.533 & 30 \\
    \cmidrule{2-10}
    & \multirow{5}{3.5em}{\centering 4-shots} & Random & - & 0.592 & 0.576 & 0 & 0.606 & 0.559 & 0 \\
        & & TF-IDF & Text + Cap. & 0.590 & 0.581 & 0 & 0.613 & 0.590 & 0 \\   
    & & TF-IDF & Rationale & 0.594 & 0.585 & 0 & 0.618 & 0.593 & 0 \\ 
    & & BM-25 & Text + Cap. & \underline{0.602} & \underline{0.590} & 0 & 0.619 & 0.594 & 0 \\    
    & & BM-25 & Rationale & 0.588 & 0.575 & 0 & \underline{0.635} & \underline{0.608} & 0 \\
    \cmidrule{2-10}
    & \multirow{5}{3.5em}{\centering 8-shots} & Random & - & 0.576 & 0.547 & 0 & 0.597 & 0.537 & 0 \\
        & & TF-IDF & Text + Cap. & 0.592 & 0.582 & 0 & 0.603 & \textbf{\underline{0.627}} & 0 \\   
    & & TF-IDF & Rationale & 0.594 & 0.581 & 0 & 0.634 & 0.611 & 0 \\ 
    & & BM-25 & Text + Cap. & \textbf{\underline{0.612}} & \textbf{\underline{0.599}} & 0 & \underline{0.636} & 0.611 & 0 \\    
    & & BM-25 & Rationale & 0.598 & 0.584 & 0 & 0.619 & 0.589 & 0 \\
    \cmidrule{2-10}
    & \multirow{5}{3.5em}{\centering 16-shots} & Random & - & 0.576 & 0.547 & 0 & 0.583 & 0.514 & 0 \\
        & & TF-IDF & Text + Cap. & 0.598 & 0.585 & 0 & 0.636 & 0.610 & 0 \\   
    & & TF-IDF & Rationale & 0.590 & 0.577 & 0 & 0.633 & 0.608 & 0 \\ 
    & & BM-25 & Text + Cap. & \underline{0.608} & \underline{0.596} & 0 & \textbf{\underline{0.644}} & \underline{0.623} & 0 \\    
    & & BM-25 & Rationale & 0.596 & 0.578 & 0 & 0.622 & 0.594 & 0 \\
    \bottomrule
  \end{tabular}
  \caption{Comparison of zero-shot and few-shot in-context learning experiment results with \textit{LatentHatred} support set across different demonstration sampling (Dem. Sampl.) strategies. \underline{Underlined} represent the best results within a dataset for the given model and given few-shot setting, \textbf{bold} indicate the best results within a dataset for a given model across all few-shot settings and \textcolor{red}{red} denote few-shot in-context learning results below zero-shot performance.}
\end{table*} 

\begin{table*}[h!]
  \small
  \centering
  \begin{tabular}{cccccccccc}
    \toprule
    & & & & \multicolumn{3}{c}{\textbf{FHM}} & \multicolumn{3}{c}{\textbf{MAMI}} \\
    \cmidrule(lr){5-7} \cmidrule(lr){8-10}
    \textbf{Model} & \textbf{\# Shots} & \textbf{Dem. Samp.} & \textbf{Matching} & \textbf{Acc.} & \textbf{F1} & \textbf{\# Invalids} & \textbf{Acc.} & \textbf{F1} & \textbf{\# Invalids} \\
    \midrule
    \multirow{16}{3.5em}{\centering LLaVA-7B} & 0-shot & - & - & 0.512 & 0.509 & 27 & 0.553 & 0.533 & 30 \\
    \cmidrule{2-10}
    & \multirow{5}{3.5em}{\centering 4-shots} & Random & - & \textbf{\underline{0.596}} & \textbf{\underline{0.576 }} & 0 & 0.591 & 0.547 & 0 \\
        & & TF-IDF & Text + Cap. & 0.578 & 0.554 & 0 & 0.611 & 0.581 & 0 \\   
    & & TF-IDF & Rationale & 0.594 & 0.571 & 0 & 0.621 & 0.600 & 0 \\ 
    & & BM-25 & Text + Cap. & 0.576 & 0.551 & 0 & 0.626 & 0.599 & 0 \\    
    & & BM-25 & Rationale & 0.570 & 0.557 & 0 & \underline{0.634} & \underline{0.610} & 0 \\
    \cmidrule{2-10}
    & \multirow{5}{3.5em}{\centering 8-shots} & Random & - & \underline{0.594} & \underline{0.575} & 0 & 0.600 & 0.556 & 0 \\
        & & TF-IDF & Text + Cap. & 0.572 & 0.546 & 0 & \textbf{\underline{0.638}} & \textbf{\underline{0.612}} & 0 \\   
    & & TF-IDF & Rationale & 0.584 & 0.568 & 0 & 0.637 & 0.613 & 0 \\ 
    & & BM-25 & Text + Cap. & 0.568 & 0.544 & 0 & 0.626 & 0.596 & 0 \\    
    & & BM-25 & Rationale & 0.584 & 0.573 & 0 & 0.635 & 0.616 & 0 \\
    \cmidrule{2-10}
    & \multirow{5}{3.5em}{\centering 16-shots} & Random & - & \red{0.378} & \red{0.362} & \red{183} & \red{0.376} & \red{0.345} & \red{374} \\
        & & TF-IDF & Text + Cap. & \red{\underline{0.426}} & \red{\underline{0.393}} & \red{113} & \red{\underline{0.509}} & \red{\underline{0.480}} & \red{208} \\   
    & & TF-IDF & Rationale & \red{0.416} & \red{0.389} & \red{150} & \red{0.486} & \red{0.447} & \red{232} \\ 
    & & BM-25 & Text + Cap. & \red{0.420} & \red{0.395} & \red{146} & \red{0.453} & \red{0.422} & \red{279} \\    
    & & BM-25 & Rationale & \red{0.124} & \red{0.118} & \red{387} & \red{0.112} & \red{0.109} & \red{812} \\
    \bottomrule
  \end{tabular}
  \caption{Comparison of zero-shot and few-shot in-context learning experiment results with \textit{FHM} support set across different demonstration sampling (Dem. Sampl.) strategies. \underline{Underlined} represent the best results within a dataset for the given model and given few-shot setting, \textbf{bold} indicate the best results within a dataset for a given model across all few-shot settings and \textcolor{red}{red} denote few-shot in-context learning results below zero-shot performance.}
   \label{tab:llava-fhm-support-set-experiments}
\end{table*}

\newpage
\subsection{Llama3-8B}
\label{appendix:llama3-experiments}

\begin{table*}[!ht]
  \small
  \centering
  \begin{tabular}{cccccccccc}
    \toprule
    & & & & \multicolumn{3}{c}{\textbf{FHM}} & \multicolumn{3}{c}{\textbf{MAMI}} \\
    \cmidrule(lr){5-7} \cmidrule(lr){8-10}
    \textbf{Model} & \textbf{\# Shots} & \textbf{Dem. Samp.} & \textbf{Matching} & \textbf{Acc.} & \textbf{F1} & \textbf{\# Invalids} & \textbf{Acc.} & \textbf{F1} & \textbf{\# Invalids} \\
    \midrule
    \multirow{16}{3.5em}{\centering Llama3-8B} & 0-shot & - & - & 0.614 & 0.586 & 7 & 0.634 & 0.596 & 5 \\
    \cmidrule{2-10}
    & \multirow{5}{3.5em}{\centering 4-shots} & Random & - & \red{0.598} & \red{0.561} & \red{9} & \red{0.569} & \red{0.499} & \red{21} \\
        & & TF-IDF & Text & \red{0.592} & \red{0.558} & \red{6} & \red{0.592} & \red{0.550} & \red{15} \\   
    & & TF-IDF & Rationale & \red{0.596} & \red{0.584} & \red{8} & \red{0.607} & \red{0.588} & \red{24} \\ 
    & & BM-25 & Text & \red{0.592} & \red{0.550} & \red{15} & \red{0.628} & \red{0.606} & \red{14} \\    
    & & BM-25 & Rationale & \red{\underline{0.602}} & \red{\underline{0.589}} & \red{3} & \underline{\red{\underline{0.628}}} & {\red{\underline{0.611}}} & \red{17} \\
    \cmidrule{2-10}
    & \multirow{5}{3.5em}{\centering 8-shots} & Random & - & \red{\underline{0.612}} & \red{0.579} & \red{3} & \red{0.592} & \red{0.531} & \red{1} \\
        & & TF-IDF & Text & \red{0.600} & \red{0.571} & \red{2} & \red{0.601} & \red{0.559} & \red{2} \\   
    & & TF-IDF & Rationale & \red{0.608} & \red{\underline{0.599}} & \red{17} & \red{0.629} & \red{0.609} & \red{20} \\ 
    & & BM-25 & Text & \red{0.601} & \red{0.559} & \red{2} & \underline{0.647} & \underline{0.627} & 8 \\    
    & & BM-25 & Rationale & \red{0.576} & \red{0.564} & \red{17} & \red{0.626} & \red{0.613} & \red{24} \\
    \cmidrule{2-10}
    & \multirow{5}{3.5em}{\centering 16-shots} & Random & - & 0.620 & 0.589 & 0 & \red{0.583} & \red{0.516} & \red{0} \\
        & & TF-IDF & Text & 0.628 & 0.606 & 0 & \red{0.605} & \red{0.563} & \red{0} \\   
    & & TF-IDF & Rationale & 0.622 & 0.610 & 0 & \red{0.625} & \red{0.603} & \red{1} \\ 
    & & BM-25 & Text & \red{0.605} & \red{0.563} & \red{0} & \textbf{\underline{0.663}} & \textbf{\underline{0.647 }}& 0 \\    
    & & BM-25 & Rationale & \textbf{\underline{0.624}} & \textbf{\underline{0.611}} & 0 & 0.658 & 0.643 & 0 \\
    \bottomrule
  \end{tabular}
  \caption{Comparison of zero-shot and few-shot in-context learning experiment results with \textit{LatentHatred} support set across different demonstration sampling (Dem. Sampl.) strategies. \underline{Underlined} represent the best results within a dataset for the given model and given few-shot setting, \textbf{bold} indicate the best results within a dataset for a given model across all few-shot settings and \textcolor{red}{red} denote few-shot in-context learning results below zero-shot performance.}
\end{table*}

\begin{table*}[!ht]
  \small
  \centering
  \begin{tabular}{cccccccccc}
    \toprule
    & & & & \multicolumn{3}{c}{\textbf{FHM}} & \multicolumn{3}{c}{\textbf{MAMI}} \\
    \cmidrule(lr){5-7} \cmidrule(lr){8-10}
    \textbf{Model} & \textbf{\# Shots} & \textbf{Dem. Samp.} & \textbf{Matching} & \textbf{Acc.} & \textbf{F1} & \textbf{\# Invalids} & \textbf{Acc.} & \textbf{F1} & \textbf{\# Invalids} \\
    \midrule
    \multirow{16}{3.5em}{\centering Llama3-8B} & 0-shot & - & - & 0.614 & 0.586 & 7 & 0.634 & 0.596 & 5 \\
    \cmidrule{2-10}
    & \multirow{5}{3.5em}{\centering 4-shots} & Random & - & \red{0.598} & \red{0.568} & \red{5} & \red{0.583} & \red{0.535} & \red{10} \\
        & & TF-IDF & Text & \red{0.598} & \red{0.569} & \red{1} & \red{0.592} & \red{0.551} & \red{11} \\   
    & & TF-IDF & Rationale & \red{0.598} & \red{0.568} & \red{1} & \red{0.633} & \red{0.606} & \red{15} \\ 
    & & BM-25 & Text & \red{\underline{0.606}} & \red{0.574} & \red{2} & \red{0.602} & \red{0.564} & \red{14} \\    
    & & BM-25 & Rationale & \red{0.600} & \red{\underline{0.585}} & \red{6} & \red{\textbf{\underline{0.627}}} & \red{\textbf{\underline{0.608}}} & \red{\textbf{15}} \\
    \cmidrule{2-10}
    & \multirow{5}{3.5em}{\centering 8-shots} & Random & - & \red{0.576} & \red{0.550} & \red{40} & \red{0.525} & \red{0.487} & \red{110} \\
        & & TF-IDF & Text & \red{0.564} & \red{0.535} & \red{29} & \red{0.546} & \red{0.518} & \red{137} \\   
    & & TF-IDF & Rationale & \red{0.566} & \red{0.545} & \red{26} & \red{0.547} & \red{0.526} & \red{131} \\ 
    & & BM-25 & Text & \red{0.560} & \red{0.536} & \red{33} & \red{0.552} & \red{0.526} & \red{132} \\    
    & & BM-25 & Rationale & \red{\underline{0.592}} & \red{\underline{0.578}} & \red{28} & \red{\underline{0.599}} & \red{\underline{0.581}} & \red{82} \\
    \cmidrule{2-10}
    & \multirow{5}{3.5em}{\centering 16-shots} & Random & - & \textbf{\underline{0.632}} & \textbf{\underline{0.608}} & \textbf{8} & \red{0.600} & \red{0.565} & \red{29} \\
        & & TF-IDF & Text & \red{0.574} & \red{0.547} & \red{7} & \red{0.610} & \red{0.581} & \red{35} \\   
    & & TF-IDF & Rationale & \red{0.610} & \red{0.591} & \red{5} & \red{0.616} & \red{0.592} & \red{38} \\ 
    & & BM-25 & Text & \red{0.590} & \red{0.563} & \red{4} & \red{0.614} & \red{0.590} & \red{40} \\    
    & & BM-25 & Rationale & \red{0.610} & \red{0.600} & \red{15} & \red{\underline{0.623}} & \red{\underline{0.611}} & \red{69} \\
    \bottomrule
  \end{tabular}
    \caption{Comparison of zero-shot and few-shot in-context learning experiment results with \textit{FHM} support set across different demonstration sampling (Dem. Sampl.) strategies. \underline{Underlined} represent the best results within a dataset for the given model and given few-shot setting, \textbf{bold} indicate the best results within a dataset for a given model across all few-shot settings and \textcolor{red}{red} denote few-shot in-context learning results below zero-shot performance.}
\end{table*}
\end{document}